%% file: main.tex
\documentclass[nohyperref]{article}

\usepackage{microtype}
\usepackage{graphicx}
\usepackage{subfigure}
\usepackage{booktabs} % for professional tables

\usepackage{hyperref}

 \usepackage[accepted]{icml2023}

\usepackage{amsmath}
\usepackage{amssymb}
\usepackage{mathtools}
\usepackage{amsthm}

\usepackage[capitalize,noabbrev]{cleveref}

\theoremstyle{plain}

\theoremstyle{definition}

\theoremstyle{remark}

\usepackage[textsize=tiny]{todonotes}

\icmltitlerunning{A Coupled Flow Approach to Imitation Learning}

\input{added_packages}

\DeclareMathOperator*{\argmax}{arg\,max} %gideon
\DeclareMathOperator*{\argmin}{arg\,min} %gideon


\begin{document}

\twocolumn[
\icmltitle{A Coupled Flow Approach to Imitation Learning}



\icmlsetsymbol{equal}{*}

\begin{icmlauthorlist}
\icmlauthor{Gideon Freund}{compdep}
\icmlauthor{Elad Sarafian}{compdep}
\icmlauthor{Sarit Kraus}{compdep}

\end{icmlauthorlist}

%\icmlaffiliation{mathdep}{Department of Mathematics, Bar-Ilan University, Israel}
\icmlaffiliation{compdep}{Department of Computer Science, Bar-Ilan University, Israel}

\icmlcorrespondingauthor{Gideon Freund}{gideonfreund@gmail.com}

\icmlkeywords{Machine Learning, ICML}

\vskip 0.3in
]



\printAffiliationsAndNotice{}

%-----------here------------

\input{sections/abstract.tex}
\input{sections/introduction}

\input{sections/background}
\input{sections/related_work}

\input{sections/our_approach}

\input{sections/experiments}

\input{sections/ablation.tex}
\input{sections/conclusion}


\bibliography{References}
\bibliographystyle{icml2023}

%%%%%%%%%%%%%%%%%%%%%%%%%%%%%%%%%%%%%%%%%%%%%%%%%%%%%%%%%%%%%%%%%%%%%%%%%%%%%%%
%%%%%%%%%%%%%%%%%%%%%%%%%%%%%%%%%%%%%%%%%%%%%%%%%%%%%%%%%%%%%%%%%%%%%%%%%%%%%%%
% APPENDIX
%%%%%%%%%%%%%%%%%%%%%%%%%%%%%%%%%%%%%%%%%%%%%%%%%%%%%%%%%%%%%%%%%%%%%%%%%%%%%%%
%%%%%%%%%%%%%%%%%%%%%%%%%%%%%%%%%%%%%%%%%%%%%%%%%%%%%%%%%%%%%%%%%%%%%%%%%%%%%%%
\newpage
\appendix
\onecolumn

\input{sections/appendix.tex}

%%%%%%%%%%%%%%%%%%%%%%%%%%%%%%%%%%%%%%%%%%%%%%%%%%%%%%%%%%%%%%%%%%%%%%%%%%%%%%%
%%%%%%%%%%%%%%%%%%%%%%%%%%%%%%%%%%%%%%%%%%%%%%%%%%%%%%%%%%%%%%%%%%%%%%%%%%%%%%%

\end{document}

%% file: added_packages.tex
\usepackage{amssymb}
\usepackage{amsmath}

%% file: sections/abstract.tex
 \begin{abstract}
In reinforcement learning and imitation learning, an object of central importance is the state distribution induced by the policy. It plays a crucial role in the policy gradient theorem, and references to it—along with the related state-action distribution—can be found all across the literature. Despite its importance, the state distribution is mostly discussed indirectly and theoretically, rather than being modeled explicitly. The reason being an absence of appropriate density estimation tools. In this work, we investigate applications of a normalizing flow-based model for the aforementioned distributions. In particular, we use a pair of flows coupled through the optimality point of the Donsker-Varadhan representation of the Kullback–Leibler (KL) divergence, for distribution matching based imitation learning. Our algorithm, Coupled Flow Imitation Learning (CFIL), achieves state-of-the-art performance on benchmark tasks with a single expert trajectory and extends naturally to a variety of other settings, including the subsampled and state-only regimes.

\end{abstract}

%% file: sections/introduction.tex
\section{Introduction}
Reinforcement learning (RL) \cite{sutton2018reinforcement} concerns the optimization of an agent's behavior in an environment. Its characterizing difficulties of exploration vs exploitation and credit assignment, stem from its typical incarnations where the agent must learn from sparse feedback. In order to provoke desired behavior, one may need to craft a sophisticated reward function or provide demonstrations for the agent to imitate. Imitation Learning (IL) deals precisely with the latter: learning from expert demonstrations. Although RL and IL share the same ultimate goal of producing a good policy, they differ fundamentally in that RL is guided by the environment’s feedback, while IL is guided only by the ability of the agent to reproduce the expert’s behavior.

Central to both RL and IL are the state and state-action distributions induced by the policy. Their importance cannot be overstated, with the state distribution forming the basis of policy gradient methods through the policy gradient theorem \cite{sutton2000policy}, and the state-action distribution being core to the common distribution matching formulation of IL \cite{ke2020imitation, ghasemipour2020divergence}. They are also foundational to other applications, like curiosity-based exploration \cite{pathak2017curiosity}, constrained RL \cite{qin2021density} and Batch RL \cite{fujimoto2019off}, some of which have recently been unified under the umbrella of convex RL \cite{zhang2020variational,zahavy2021reward,mutti2022challenging}, which studies objectives that are convex functions of the state distribution.  

% maybe cut the line short (remove who do attempt to model it)
Despite their ubiquity across the literature, explicit modeling of the distributions is scarce. Instead, they mostly find use as a theoretical tool for derivations. This is, of course, barring some approaches that do attempt to model them \cite{hazan2019provably,qin2021density,lee2019efficient,kim2021imitation} or their ratios \cite{nachum2019dualdice,liu2018breaking,gangwani2020harnessing}, further discussion of which is delegated to the related work. The reason for this lack of modeling is due to the difficulty of density estimation, especially in the case of complex and high-dimensional distributions. This is where normalizing flows (NF) \cite{dinh2014nice,dinh2016density, papamakarios2017masked}, a recent approach to density estimation, will be of service. 

%perhaps

%[should motivate why imitation is a natual place for such a model]
We believe there is no shortage of applications for such modeling, but we focus on imitation learning, a quite natural and suitable place, given its modern formulation as state-action distribution matching \cite{ghasemipour2020divergence}. 

Many approaches to distribution matching based imitation have been presented \cite{ho2016generative,fu2017learning,kostrikov2018discriminator,ke2020imitation, ghasemipour2020divergence, kostrikov2019imitation,sun2021softdice,kim2021imitation, dadashi2020primal,schroecker2020manipulating}. The common theme for such methods begins with the selection of a divergence, followed by the development of a unique approach. This may involve a direct attack on the objective by reformulating it \cite{kostrikov2019imitation}, or by derivation of a surrogate objective \cite{zhu2020off, dadashi2020primal, kim2021imitation}, with some utilizing mechanisms such as an inverse action model \cite{zhu2020off} and focusing on learning from states alone (a setting our approach naturally lends to). Other methods first derive estimates for the gradient of the state distribution with respect to the policy's parameters \cite{schroecker2020manipulating}, while some devise unifying algorithms and frameworks encompassing previous approaches \cite{ke2020imitation, ghasemipour2020divergence}.

The most popular divergence of choice is the reverse KL, which some favor due to its mode-seeking behavior \cite{ghasemipour2020divergence}. Others attempt to get the best of both worlds, combining both mode-seeking and mode-covering elements \cite{zhu2020off}. A priori, it is difficult to say which choice of divergence is advantageous, it's more about the ensuing approach to its minimization.

In this work, we propose a unique approach to distribution matching based imitation, by coupling a pair of flows through the optimality point of the Donsker-Varadhan \cite{donsker1976asymptotic} representation of the KL. More specifically, by noting this point occurs at the log distribution ratio, while the IL objective with the reverse KL can be seen as an RL problem with the ratio inverted. We propose setting the point of optimality as the difference of two normalizing flows, then training in an alternating fashion akin to other adversarial IL methods \cite{ho2016generative,kostrikov2018discriminator, ghasemipour2020divergence, kostrikov2019imitation,sun2021softdice}. This method proves far more accurate than estimating the log distribution ratio by naively training a pair of flows independently. We show this in part by analyzing their respective BC graphs: a simple tool we present for gauging how well a proposed estimator captures the expert's behavior. While most IL works neglect analysis of their learned reward function, we think this can be a potential guiding tool for future IL researchers.

Our resulting algorithm, Coupled Flow Imitation Learning (CFIL) shows strong performance on standard benchmark tasks, while extending naturally to the subsampled and state-only regimes. In the state-only regime in particular, CFIL exhibits significant advantage over prior state-of-the-art work, despite the competition being specifically designed for that domain. This work also aims to inspire more research incorporating explicit modeling of the state-action distribution.

%% file: sections/background.tex
\section{Background}

\subsection{Markov Decision Process}
Environments in RL are typically expressed as a Markov Decision process (MDP). An MDP is a tuple $(S,A,P,R,p_0,\gamma)$, where $S$ is a set of states termed the state space, $A$ is a set of actions termed the action space, $P: S \times A \times S \to [0,\infty]$ is a transition density function describing the environment's Markovian dynamics, $R: S\times A \times S \to \mathbb{R}$ is a reward function, $p_0$ is an initial state distribution, and $\gamma \in [0,1)$ is a discount factor. \par
A policy $\pi: S \to A$ dictates an agent's actions when roaming the environment, and the goal in an MDP is to find the optimal one, denoted $\pi^*$. The standard optimality criterion for a policy is the expected discounted reward:
$J(\pi,r) = \mathbb{E}_{\tau \sim \pi}\left[\sum_{t=0}^{\infty}\gamma^t r_t \right]$
where $\tau \sim \pi$ symbolizes the trajectory distribution induced by the policy $\pi$, and $r_t$ is the reward at time $t$ along a trajectory. Each policy has an associated value function and Q-function:
$V^\pi(s) = \mathbb{E}_\pi\left[\sum_{t=0}^{\infty}\gamma^t r_t | s_0 = s\right] \text{\, and \,}
Q^\pi(s,a)= \mathbb{E}_\pi\left[\sum_{t=0}^{\infty}\gamma^t r_t | s_0 = s, a_0 = a\right],$
where the value function represents the expected discounted reward of following $\pi$ from state $s$, while the Q-function represents the expected discounted reward of following $\pi$ after taking action $a$ from state $s$. 

Another object induced by the policy is the improper discounted state distribution $d^\gamma_{\pi}(s) = \sum_{t=0}^{\infty}\gamma^t Pr(s_t = s | s_0 \sim p_0),$ as well as its undiscounted counterpart, $d_{\pi}(s) = \sum_{t=0}^{\infty}Pr(s_t = s | s_0 \sim p_0),$ in which we take greater interest for reasons described in Appendix \ref{appendix:undiscounted}. $d_{\pi}(s)$ is called the state distribution and is of central interest to this work. Closely related to $d_\pi(s)$ is the state-action distribution $p_\pi(s,a) = \pi(a|s)d_\pi(s)$, which is also of interest to this work.

\subsection{Normalizing Flows}\label{background:flows}
Normalizing Flows (NF) \cite{dinh2014nice,dinh2016density} are exact-likelihood generative models that use a simple base distribution $p_Z(z)$ and a sequence of invertible and differentiable transformations $f_{\theta} = f_k \circ f_{k-1} \circ \dots \circ f_1$ to model an arbitrary target distribution $p_X(x)$ as $z = f_{\theta}(x)$.
\footnote{More often presented as $x = g(z)$ \cite{papamakarios2021normalizing,kobyzev2020normalizing}, which is equivalent with $f=g^{-1}$, due to their invertibility. Under this model, Equation \eqref{eq:NF} would look similar with only the "$+$" substituted by a "$-$".}
This means the target log-likelihood can be written using the change of variables formula:
\begin{equation}\label{eq:NF}
\log p_X(x) =
\log  p_Z(z) + \sum_{i=1}^{k}{\log \left|\text{det}\frac{df_i}{df_{i-1}}\right|}
\end{equation}
allowing for parameters $\theta$ to be trained using maximum likelihood. With a trained flow at hand, generation is performed by sampling from the base distribution and applying $f_{\theta}^{-1}$, and density estimation can be done by evaluating the RHS of \eqref{eq:NF}. RealNVP \cite{dinh2016density}, which is of particular interest to us, uses coupling layers (not to be confused with our coupled approach) to construct $f_{\theta}$. Masked Autoregressive Flow (MAF) \cite{papamakarios2017masked} is a generalization of RealNVP which substitutes its coupling layers with autoregressive ones. While improving expressiveness, this impedes single pass sampling, but still allows efficient density evaluation—our main concern—using masked autoencoders \cite{germain2015made}. RealNVP and MAF are reviewed more thoroughly in Appendix \ref{appendix:flows}.

Although much research \cite{kingma2018glow,huang2018neural,durkan2019neural,ho2019flow++} has gone into improving the capacity of MAF, due to its simplicity, efficiency and adequacy for our task, it is the architecture used in this work to model the state and state-action distributions $d_\pi(s)$ and $p_\pi(s,a)$.

\subsection{Imitation Learning}
Imitation Learning (IL) concerns the optimization of an agent’s behavior in an environment, given expert demonstrations. Perhaps the simplest approach to imitation, behavioral cloning (BC), performs supervised regression or maximum-likelihood on given expert state-action pairs $\{(s_e,a_e)\}_{t=1}^N$: \begin{equation}\label{eq:bc_objective} {\min_\pi \sum_{t=0}^N Loss(\pi(s_t),a_t)\quad \text{or} \quad \max_\pi \sum_{t=0}^N \log \pi(a_t|s_t)} \end{equation} for deterministic and stochastic policies, respectively. BC suffers from compounding errors and distributional shift, wherein unfamiliar states cause the policy to misstep, leading to even less familiar states and an eventual complete departure from the expert trajectory \cite{ross2011reduction}.

Recent distribution matching (DM) approaches \cite{ho2016generative,kostrikov2018discriminator,kostrikov2019imitation,kim2021imitation} successfully overcome these issues. One common formulation \cite{ke2020imitation, ghasemipour2020divergence} encompassing most of these methods, views DM as an attempt to match the agent's state-action distribution $p_\pi$, with the expert's $p_{e}$, by minimizing some f-divergence\footnote{Other distances, such as the Wasserstein metric have also been used \cite{sun2021softdice,dadashi2020primal}.} $D_f$: 
\begin{equation}\label{eq:Matching}
\argmin_{\pi} \, D_f(p_\pi||p_e).
\end{equation} 
These methods hinge on the one-to-one relationship between a policy and its state-action distribution and have shown significant improvement over BC, particularly when few expert trajectories are available \cite{ghasemipour2020divergence} or expert trajectories are subsampled \cite{li2022rethinking}. 

%% file: sections/related_work.tex
\section{Related Work}

Attempts at modeling the state distribution have been made for various applications and countless distribution matching approaches to imitation learning exist, with varying degrees of relevancy to our own. The most pertinent works—those intersecting both the former and the latter—are reviewed in most detail towards the end of this section.

Some general attempts to model the state distribution include \cite{hazan2019provably} who use discretization and kernel density estimates (KDE) for curiosity-based exploration, while \cite{lee2019efficient} uses variational autoencoders (VAE) in the same context. VAE's have also been used by \cite{islam2019off} to model $d_\pi(s)$ for constraining the distribution shift in off-policy RL algorithms, and KDE's have also been used by \cite{qin2021density} for density constrained reinforcement learning.

Approaches to distribution matching-based imitation include GAIL \cite{ho2016generative} who uses a GAN-like \cite{goodfellow2014generative} objective to minimize the JS divergence. Originally born out of max-entropy inverse reinforcement learning \cite{ziebart2008maximum}, GAIL paved the way for later works such as AIRL \cite{fu2017learning} which modified GAIL's objective to allow reward recovery; DAC \cite{kostrikov2018discriminator}, an off-policy extension of GAIL also addressing reward bias; and general f-divergence approaches like \cite{ke2020imitation, ghasemipour2020divergence}. Other works include the DICE \cite{nachum2019dualdice} family comprised of ValueDICE \cite{kostrikov2019imitation} and its successors SoftDICE \cite{sun2021softdice}, SparseDICE \cite{camacho2021sparsedice} and DemoDICE \cite{kim2021demodice}. Using the Donsker-Varadhan representation along with a change of variables, ValueDICE derives a fully off-policy objective from the reverse KL, completely drubbing BC in the offline regime. Its successors are of tangential relevancy here, each augmenting it for a different domain: SoftDICE makes various amendments notably using the Wasserstein metric (also used by PWIL \cite{dadashi2020primal}), while SparseDice adds a host of regularizers to extend ValueDICE to subsampled trajectories and DemoDICE focuses on imitation with supplementary imperfect demonstrations. We note the apparent advantages of the DICE family over BC in the fully offline regime have recently been questioned \cite{li2022rethinking}.

Our exploitation of Donsker-Varadhan was certainly inspired by ValueDICE and is somewhat reminiscent of MINE \cite{belghazi2018mine} who utilize it to estimate mutual information with a regular neural estimator. Their context and precise method however, differ substantially.
Another work of relation is the often overlooked \cite{schroecker2020manipulating} which includes three approaches: SAIL, GPRIL and VDI \cite{ schroecker2017state, schroecker2019generative, schroecker2020universal}. All three attack the IL problem via the forward KL. Essentially, the forward KL is broken into a behavioral cloning term as well as a state distribution term, and optimizing the objective then requires an estimate of $\nabla_\theta \log d_{\pi_\theta}(s)$. SAIL, GPRIL and VDI each propose a unique way of estimating this gradient, with GPRIL and VDI incorporating flows. \cite{chang2022flow} also utilize flows, proposing a state-only IL approach that models the state-next-state transition using conditional flows, but require dozens of expert trajectories to find success. 

Finally, the most similar work is NDI \cite{kim2021imitation} who rewrite the reverse KL as $-D_{KL}(p_\pi||p_{e}) = \mathbb{E}_{p_\pi}\left[\log p_e - \log p_\pi \right] = J(\pi, r{=}\log p_e) + \mathcal{H}(p_\pi).$ That is, RL with rewards $\log p_e$, along with a state-action entropy term. They continue by deriving a lower bound on $\mathcal{H}(p_\pi)$, termed the SAELBO. NDI+MADE is then proposed, where in a first phase $\log p_e$ is estimated using flows followed by a second phase of optimization.

NDI has many limitations and differs dramatically from our approach. NDI claims to be non-adversarial, while true, it is only due to their loose bound. Moreover, this bound was proven superfluous in their ablation where setting $\lambda_f {=} 0$ showed no reduction in performance. 
On top of that, we far outperform them, our evaluation is far more comprehensive and although we both use flows, the method and employment differ significantly in CFIL given the coupling of a pair with Donsker-Varadhan.

Pausing the deturpation, each of these methods above has strong merits and heavily inspired our own: We adopted the use of MAF over RealNVP from NDI+MADE (our work began in ignorance of NDI); and it was studying ValueDICE that inspired our direction to exploit the Donsker-Varadhan representation of the KL. The way in which we do however, is unique, and all the approaches above are distinct from our own.

%% file: sections/our_approach.tex
\section{Our Approach}\label{sec:our_approach}
%[maybe prefacing line about IL] towards our IL approach
We begin with the reverse KL as our divergence of choice, since, noting as others have \cite{kostrikov2019imitation,hazan2019provably,kim2021imitation,camacho2021sparsedice}, its minimization may be viewed as an RL problem with rewards being the log distribution ratios:
\begin{align}
    \argmin_{\pi}\,D_{KL}(p_\pi||p_e) =& \argmax_{\pi}\,\mathbb{E}_{p_\pi(s,a)}\left[\log \frac{p_e(s,a)}{p_\pi(s,a)} \right] \nonumber \\=& 
    \argmax_{\pi}J(\pi, r{=}\log \frac{p_e}{p_\pi}).\label{eq:rKL}
\end{align}
Thus permitting the use of any RL algorithm for solving the IL objective, provided one has an appropriate estimate of the ratio.

%briefly
Before continuing however, we first motivate our coupled approach for such estimation, by illustrating the failure of what is perhaps a more natural next step: using two independent density estimators—say flows—for each of the densities $p_e$ and $p_\pi$ directly. Practically, this would mean alternating between learning the flows %(like Equation \ref{eq:regularization} below)
and using their log-ratio as reward in an RL algorithm. The table in Section \ref{section:ablation} concisely showcases a clear failure of this approach on all the standard benchmarks we later evaluate with.

\begin{figure}[t]
\vskip 0.1in
\centering
\includegraphics[width=0.95\columnwidth]{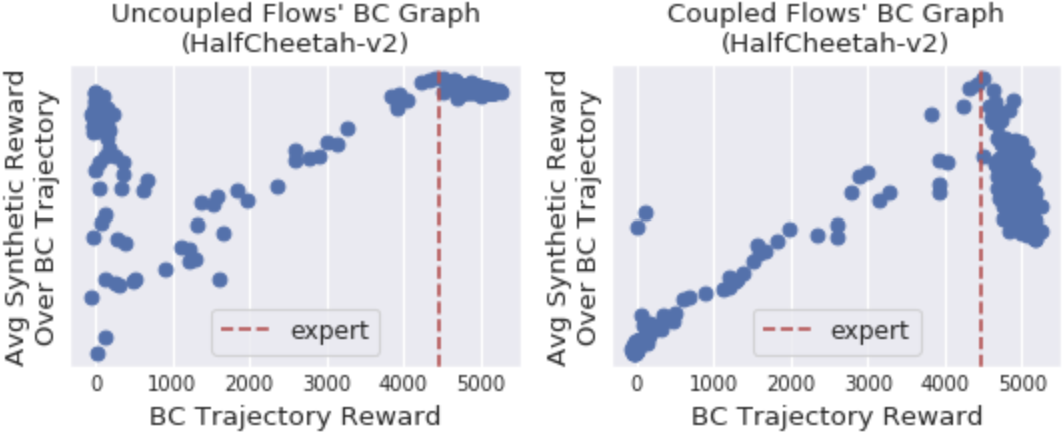} % Reduce the figure size so that it is slightly narrower than the column. Don't use precise values for figure width.This setup will avoid overfull boxes.
\caption{Left: The BC graph of an uncoupled flow for the HalfCheetah-v2 environment. Right: The BC graph of a coupled flow for the HalfCheetah-v2 environment. BC graphs for an estimator are generated by updating the estimator analogously to an RL run using $N$ saved BC rollouts. This yields $N$ estimators corresponding to $N$ intermediate BC agents. The BC graph is then, for all $i$, the scatter of the $i$’th estimator’s evaluation of an $i$’th BC agent's trajectory against its true environment reward.}
\label{fig:ind_flows_failure_and_bc}
\vskip -0.1in
\end{figure}

The failure can be further understood through analyzing what we term the BC graph corresponding to an estimator of the log distribution ratio. That is, we first train a behavioral cloning agent once on sufficient expert trajectories, while rolling it out every few iterations. We then train the estimator analogously to an RL run, using a single expert trajectory along with the $N$ saved BC rollouts. This quick process yields $N$ estimators corresponding to $N$ intermediate BC agents. The BC graph is then, for all $i$, the $i$’th estimator’s evaluation of an $i$’th BC agent's trajectory, scattered against the true environment reward of that same trajectory. \footnote{Meaningfulness of the BC graph depends on how erratic the training was and how iteration number, loss and true environment reward line up during training. Our choice of Cheetah for illustration is motivated due to all these measures mostly aligning.} 
Intuitively, a non increasing graph, means a potential RL learner with an analogously trained reward may struggle to overcome those misleading behaviors ranked high by the synthetic reward, thus preventing them from reaching expert level. In practice of course, one would want some form of approximate monotonicity or upward trend. Though importantly, a BC graph's monotonicity by no means implies the success of an RL learner with the correspondingly constructed reward. This is more than a theoretical idiosyncrasy: Many estimators will emit a perfect BC graph while completely failing all attempts in RL (see Appendix \ref{section:appendix_regular_net_bc}). Only the reverse is true: its complete lack of an upward trend will usually imply an agent's failure.

Loosely formalized, given BC's objective in Equation \ref{eq:bc_objective} and assuming a stationary (time independent) reward function $r$, a monotonically increasing BC graph essentially\footnote{Once again assumes alignment of environment reward with loss.} means that for all policies $ \pi_1, \pi_2$: $ J(\pi_1,r) > J(\pi_2,r)$ if $ \mathbb{E}_{p_e}[\log \pi_1(a|s)]  > \mathbb{E}_{p_e}[\log \pi_2(a|s)] $. Thus, further assuming continuity, a non monotonically increasing graph implies it either monotonically decreases or the existence of a local maximum. In both cases an RL learner with objective $\argmax_\pi{J(\pi,r)}$ may converge on a policy with suboptimal BC loss. Since only at optimality BC recovers the expert policy, this would guarantee the agent will not meet its truly intended goal of expert mimicry. 
Of course, in reality, RL can overcome a certain lack of an upward trend. Moreover, the rewards are neither stationary nor identical between the BC and RL runs, only analogously constructed, so such graphs are only loosely representative. Nonetheless, we find they can be highly insightful.

As Figure \ref{fig:ind_flows_failure_and_bc} suggests, the independently trained flows' BC graph is quite lacking. An agent would have no incentive according to the synthetic reward to make any progress, which is precisely what occurs as the table in Section \ref{section:ablation} demonstrates. This poor BC graph is due in part to each flow being evaluated on data completely out of its distribution (OOD), which flows are known to struggle with \cite{kirichenko2020normalizing}. Since the two flows estimates lack true meaning when evaluated on each others data, we need to tie them together somehow: They must be \textit{coupled}.

To perform our coupling, we employ the Donsker-Varadhan \cite{donsker1976asymptotic} form of the KL divergence:
%wanted to align but too big... how can it be done?
\begin{align}
D_{KL}&(p_\pi||p_{e}) =
\nonumber \\
&\sup_{x: S \times A \to \mathbb{R}} \mathbb{E}_{p_\pi(s,a)}\left[x(s,a)\right] - \log \mathbb{E}_{p_e(s,a)}\left[e^{x(s,a)} \right].\label{eq:Donsker}
\end{align}
 %note the "&" in the KL helps properly align equations
In the above, optimality occurs with  $x^*= \log \frac{p_\pi}{p_e} + C$ for any $C\in \mathbb{R}$ \cite{kostrikov2019imitation, gangwani2020harnessing,belghazi2018mine}. Thus after computing $x^*$, one recovers the log distribution ratio by simple negation, enabling use of $-x^*$ as reward in an RL algorithm to optimize our IL objective. This leads directly to our proposed approach for estimating the log distribution ratio, by coupling two flows through $x(s,a)$. That is, instead of training two flows independently, we propose to do so through maximization of Equation \ref{eq:Donsker}. More specifically, we inject the following inductive bias, modeling $x$ as $x_{\psi,\phi}(s,a) = \log p_\psi(s,a) - \log q_\phi(s,a),$ where  $p_\psi$ and $q_\phi$ are normalizing flows. %[so by maximizing \ref{eq:Donsker} one recovers x...] 

This coupling guarantees more meaningful values when the flows are evaluated on each others data, since it has already occurred during the maximization phase, hence sidestepping the OOD issue described earlier. Figure \ref{fig:ind_flows_failure_and_bc} illustrates this advantage. The right shows the coupled flows' BC graph, clearly remedying the issue with their uncoupled counterparts: A potential learner will now have the proper incentive to reproduce the expert's behavior. 

The drop in synthetic reward (i.e. $-x^*$) towards the end of the BC graph may seem daunting, but it actually expresses how well our estimator captures the expert's behavior: The drop occurs precisely beyond expert level, where the agent, while good in the environment, diverges from the true expert's behavior.\footnote{ This both illustrates the tendency of BC to overfit \cite{li2022rethinking}, while also raising concerns about the tendency for IL papers to report a table showing slight advantage in asymptotic reward as being meaningful. In truth, once at expert level, an advantage should not be claimed for slightly higher performance, unless the stated goal of the work is to do so, but that would no longer be imitation.}

Given this improved estimator, our full IL objective can then be written as:
\begin{equation}\label{eq:maxmin} 
    \argmax_{\pi} \min_{p_\psi, q_\phi}
    \log \mathbb{E}_{p_e(s,a)}\left[e^{\log \frac{p_\psi}{q_\phi}} \right] - \mathbb{E}_{p_\pi(s,a)}\left[\log \frac{p_\psi}{q_\phi}\right].
\end{equation}
As is commonplace in adversarial like approaches \cite{goodfellow2014generative, ho2016generative,kostrikov2019imitation}, the max-min objective above is trained in an alternating fashion, switching between learning $x$ and using $-x$ as reward in an RL algorithm. Moreover, we find it useful to use a squashing function on $x$, avoiding a potentially exploding KL, due to a lack of common support (subtly different then the earlier issue of OOD). 

%maybe mention the shift as well: thanks to RL's invariance to constant reward shift $\argmax_\pi J(\pi, r) = \argmax_\pi J(\pi, r+C)$

Our approach still enables training each flow independently along the way. We call this flow regularization and importantly, our method succeeds without such regularization. More specifically, for expert and agent batches of size $M$, this
regularization involves incorporating the following additional loss function into the minimization step of Equation \ref{eq:maxmin}: 
\begin{equation}\label{eq:regularization}
\mathcal{L} = - \frac{1}{M} \sum_{i=1}^M \log  q_\phi(s_e^i,a_e^i) + \log  p_\psi(s^i,a^i),
\end{equation}
with $\mathcal{L}$ to be weighted by a coefficient $\alpha$.

Noting that training flows is a delicate process, our approach further benefits from—though again does not require—use of a smoothing akin to the dequantization used when training normalizing flows on discrete data \cite{papamakarios2017masked}. More specifically, since our input is unnormalized, we smooth each dimension with uniform noise scaled to its value. That is, if $(s,a)$ is the vector to be smoothed, we sample uniform noise with dimension $dim((s,a))$, multiply them element-wise and add that to the original vector:
\begin{equation}
    {(s,a) \mathrel{{+}{=}} \beta \cdot (s,a) \odot u, \hspace{0.5em} u \sim Uniform(-\frac{1}{2},\frac{1}{2})^{dim((s,a))}},
\end{equation}
with weight $\beta$ controlling the smoothing level. Note if regularization is also present, smoothing still applies within the additional loss $\mathcal{L}$.

Finally, combining  all the above is our resulting algorithm, Coupled Flow Imitation Learning (CFIL). It is summarized in Algorithm \ref{alg:CFIL}, with only the number of batches per density update omitted. 
%algorithm notes:
%
\begin{algorithm}[tb]
\caption{CFIL}
\label{alg:CFIL}
\textbf{Input}: Expert demos $\mathcal{R}_E = \{(s_e,a_e)\}_{t=1}^N$; parameterized flow pair $p_\psi,q_\phi$; off-policy RL algorithm $\mathcal{A}$; density update rate $k$; squashing function $\sigma$; regularization and smoothing coefficients $\alpha, \beta$.\\
\textbf{Define}: $x_{\psi,\phi} = \sigma(\log p_\psi - \log q_\phi)$
\begin{algorithmic}[1] %[1] enables line numbers
\FOR{timestep $t=0,1,\dots,$}
\STATE Take a step in $\mathcal{A}$ with reward $r=-x_{\psi,\phi}$, while filling agent buffer $\mathcal{R}_A$ and potentially updating the policy and value networks according to $\mathcal{A}$'s settings.
\IF {$t\mod k =0$}
\STATE Sample expert and agent batches:
\STATE $\{(s_e^t,a_e^t)\}_{t=1}^M \sim \mathcal{R}_E$ and  $\{(s^t,a^t)\}_{t=1}^M \sim \mathcal{R}_A$
\IF{smooth}
\STATE ${(s,a) \mathrel{{+}{=}} \beta \cdot (s,a) \odot u, \hspace{0.5em} u \sim U(-\frac{1}{2},\frac{1}{2})^{dim((s,a))}}$ 
\ENDIF
\STATE Compute loss: %[possibly reference equation in paper]:
\STATE $\mathcal{J} = \log \frac{1}{M} \sum_{i=1}^M e^{x(s_e^i,a_e^i)} - \frac{1}{M} \sum_{i=1}^M x(s^i,a^i)$
\IF {flow reg}
\STATE Compute regularization loss:
\STATE $\mathcal{L} = - \frac{1}{M} \sum_{i=1}^M \log  q_\phi(s_e^i,a_e^i) + \log  p_\psi(s^i,a^i)$
\STATE $\mathcal{J} = \mathcal{J} + \alpha \mathcal{L}$
\ENDIF
\STATE Update $\psi \leftarrow \psi - \eta \nabla_{\psi}\mathcal{J}$
\STATE Update $\phi \leftarrow \phi - \eta \nabla_{\phi}\mathcal{J}$
\ENDIF
\ENDFOR

%"curly brackets turn an equation into a math atom and prevent it from breaking up"

\end{algorithmic}
\end{algorithm}
As in ValueDICE \cite{kostrikov2019imitation}, we found the bias due to the log-exp over the mini-batches did not hurt performance and was therefore left unhandled.

Another setting of interest is learning from observations (LFO) alone \cite{zhu2020off,torabi2018generative,torabi2021dealio}. That is, attempting to minimize: \begin{equation}
\argmin_{\pi} \, D_{KL}(d_\pi(s,s')||d_e(s,s')).
\end{equation}
While this objective is clearly underspecified in a non injective and deterministic MDP, in practice, recovering the expert's behavior is highly feasible \cite{zhu2020off}.
%injective meaning all actions lead to different states and deterministic meaning actions always lead to a state
%(since different actions may lead to the same state, hence multiple policies can induce identical state next-state distributions)
%so not one to one.
Seeing as none of our description above is specific to the domain of states and actions, CFIL naturally extends to LFO with no need of modification. This is in stark contrast to previous works in the LFO setting which have been highly tailored \cite{zhu2020off}. We shall demonstrate CFIL's utility in the LFO setting in the following section, where remarkably, we even find success when the flows model the single state distribution $d(s)$.

%% file: sections/experiments.tex
\section{Experiments}\label{sec:experiments}
We evaluate CFIL on the standard Mujoco benchmarks \cite{todorov2012mujoco}, first comparing it to state-of-the-art imitation methods, including ValueDICE \cite{kostrikov2019imitation} and their optimized implementation of DAC \cite{kostrikov2018discriminator}, along with a customary behavioral cloning (BC) baseline. We then move to evaluation on a variety of other settings described below. We use ValueDICE's original expert demonstrations, with exception to the Humanoid environment, for which we train our own expert, since they did not originally evaluate on it. We use ValueDICE's open-source implementation to comfortably run all three baselines. NDI \cite{kim2021imitation} would be the ideal candidate for comparison, given the similarities, however no code was available. Still, we reference some relevant results described in their paper.

For CFIL, our RL algorithm of choice is SpinningUps's \cite{SpinningUp2018} SAC \cite{haarnoja2018soft}. We leave all hyper-parameters unchanged, only reducing the start-steps down to 2000, matching that of the baselines above.
Our choice for both flows is a single layered MAF \cite{papamakarios2017masked}, amending no hyper-parameters from the following open-source implementation \cite{MAFcode}. This lack of tuning highlights the robustness of our method. 
Our density update rate is 10 batches of 100, every 1000 timesteps. We use the Adam optimizer \cite{kingma2014adam} with a learning rate of 0.001. For squashing we use $\sigma = 6 tanh(\frac{x}{15})$, while the smoothing and regularization coefficients are 0.5 and 1 respectively.

For all algorithms, we run 80 epochs, each consisting of 4000 timesteps, evaluating over 10 episodes after each.
We do this across 5 random seeds and plot means and standard deviations. The plots are smoothed by a rolling mean with window length 5. All results are using a single expert trajectory. Appendix \ref{appendix:CFIL_varying_traj} shows CFIL's performance with more expert trajectories.

%figure * helps be on one column
\begin{figure*}[t!]
\vskip 0.1in
\centering
\includegraphics[width=0.9\textwidth]{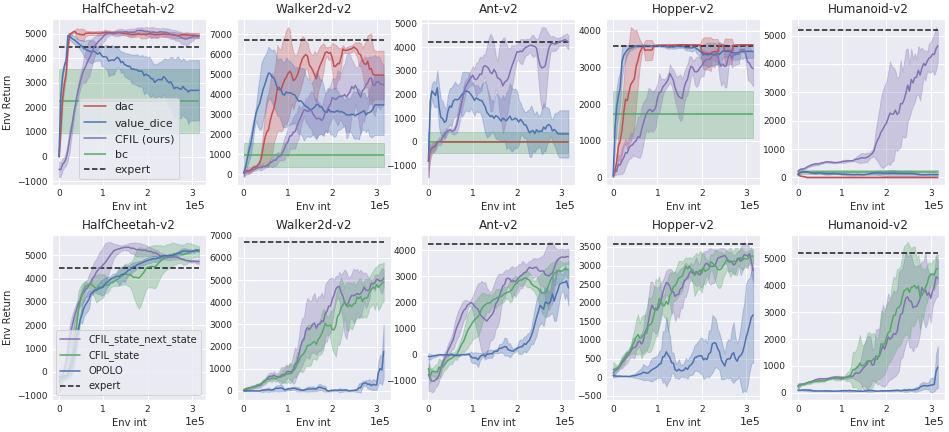} % Reduce the figure size so that it is slightly narrower than the column. Don't use precise values for figure width.This setup will avoid overfull boxes.
\caption{Top: A comparison of CFIL to ValueDICE and DAC on a single expert trajectory in the standard state-action setting. Bottom: A comparison of two versions of CFIL to OPOLO on a single expert trajectory in the LFO setting, with one version limiting itself only to single states. CFIL uses identical hyperparameters on all environments in all three incarnations, showing outstanding results, particularly in the LFO setting where it far outperforms the highly tailored competitor OPOLO.}
\label{fig:CFILcompare}
\vskip -0.1in
\end{figure*}
The top of Figure \ref{fig:CFILcompare} shows a comparison to the baselines in the standard state-action setting. Clearly, CFIL achieves expert level performance on all tasks with merely a single expert trajectory—something the baselines fail to do. CFIL either outperforms or performs similarly to the competition in terms of asymptotic performance (though slight outperformace is not meaningful once all are at expert level), with a massive advantage in the Ant and Humanoid environments.

In terms of environment interactions required until reaching expert level: Our approach slightly lags behind the optimized DAC and largely lags behind ValueDICE. That is of course, only for the tasks they succeed on. ValueDICE—where it succeeds—clearly wins in terms of interactions needed to reach expert level, though its performance then occasionally degrades as seen in Figure \ref{fig:CFILcompare}. Being semi-offline (i.e. it can be instantiated fully offline), ValueDICE is difficult to compete with in terms of environment interactions. 
However, minimizing interactions was not the objective of this work, yet still we are on par. 

Despite not being our aim, we do note that CFIL far outperforms NDI which requires an order of magnitude more interactions (see their appendix), who in turn outperform GAIL \cite{ho2016generative} by another order of magnitude. Clearly, CFIL is still very impressive in terms of environment interactions and the lag between us and DAC may simply be due to a lack of tuning of CFIL's RL component. We deliberately avoid this tuning, preferring to keep the RL algorithm a black box, since too much tuning may jeopardize robustness and extendability to other settings: We aspire for robust competitiveness rather than a tailored, minor and frail improvement.

We now turn to the state-only and subsampled regimes. Settings in which ValueDICE finds no dice: By its very nature, ValueDICE is inapplicable to the state-only scenario (see Appendix 9.8 of \cite{zhu2020off}) and extensive experiments already showed its failure when demonstrations are subsampled \cite{li2022rethinking}. SparseDICE \cite{camacho2021sparsedice} attempted to remedy this by adding a series of regularizers. However, they require at least 10 demonstrations, each subsampled at a rate of 0.1, not nearly the sparsity of the settings we next describe.

First in the state-only regime, we compare against OPOLO \cite{zhu2020off}, a state-of-the-art LFO method. We avoid comparison with on-policy LFO methods, like GAIfO \cite{torabi2018generative}, due to the sheer number of interactions they require. We use OPOLO's open-source implementation, with the same setup described previously, once again using only a single expert trajectory. Importantly, CFIL uses identical hyperparameters to the previous setting.

The bottom of Figure \ref{fig:CFILcompare} shows two versions of CFIL compared with OPOLO. One estimating the state-next-state distribution ratio, while the other limiting itself to the single state distribution. As can be seen, both incarnations of CFIL once again achieve expert level performance on all environments. This is in sharp contrast to OPOLO which, despite being highly tailored towards the LFO setting, employing inverse action models, discriminators and a host of practical considerations, struggles immensely when provided with a single expert trajectory (they originally report success with 4 trajectories). All the while, CFIL requires no amendment whatsoever, extending simply and elegantly, while distinctly outdoing OPOLO on every imitation learning metric. CFIL's results in the LFO setting are nothing short of remarkable. We effectively set a new state-of-the-art for learning from observations alone, while our method was not originally designed for it.

We finally turn to the subsampled regime. Comparing again to DAC with a similar setup as before. Our comparison here involves four different subsampling rates, sampling every 10, 20, 50 and 100'th transition. Since we are still working with a single expert trajectory, the final rate implies learning from 10 state-action pairs alone. For this setting, we found the hyperparameters used above did not extend well enough. Specifically, the use of flow regularization seemed to hurt performance in some environments. We therefore drop the regularization to 0, leaving the smoothing at 0.5 and amending the squasher to be $\sigma = 3 tanh(\frac{x}{10})$. We use these parameters for all environments in all four new settings, with exception to cheetah who, when subsampled, actually struggles without the flow regularization.

\begin{figure}[t]
\vskip 0.1in
\centering
\includegraphics[width=0.9\columnwidth]{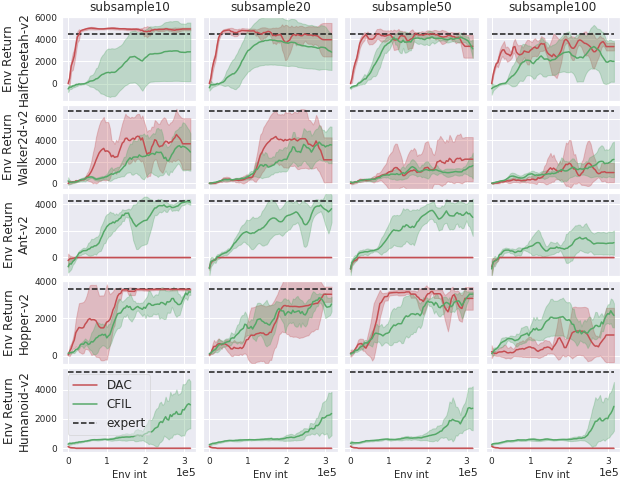} % Reduce the figure size so that it is slightly narrower than the column. Don't use precise values for figure width.This setup will avoid overfull boxes.
\caption{A comparison of CFIL and DAC on four subsampling rates with a single expert trajectory. Subsample$N$ refers to sampling every $N$'th transition in the trajectory.}
\label{fig:subsample}
\vskip -0.1in
\end{figure}

Figure \ref{fig:subsample} shows the comparison in the subsampled setting, with CFIL once again demonstrating stellar performance. While DAC has the advantage in the Cheetah environment, where single crashing seeds significantly reduced the apparent performance of CFIL, CFIL still generally outperforms it, remarkably able to recover the expert behavior when only provided with a measly 10 state-action pairs.

%% file: sections/ablation.tex
\section{Ablation}\label{section:ablation}
We now concisely ablate various aspects of our approach. We first put into question the need for our squasher, our coupling and our inductive bias, by comparing to NoSquash; IndFlow and IndFlowNS; and RegularNet. NoSquash is CFIL stripped of the squasher. IndFlow and IndFlowNS refer to learning the log distribution ratio directly using independent flows (as in section \ref{sec:our_approach}), with and without a squasher, respectively. RegularNet alters CFIL's inductive bias by setting $x$ to a regular MLP, completely avoiding the use of flows (reminiscent of MINE \cite{belghazi2018mine}). Along with these we also run a numerator only approach termed Numerator, which simply involves direct learning of the expert's distribution with a single flow, then using it alone as reward in an RL algorithm: $J(\pi, r{=}\log p_e)$  (akin to NDI \cite{kim2021imitation} with $\lambda_f=0$). 

We run these with our usual setup of a single expert trajectory and compute their overall score as the average normalized asymptotic reward over all $25$ seeds ($5$ for each environment). Table \ref{table:ablation_regnet} summarizes these results, showing all the CFIL alternatives fail, demonstrating the necessity of its components.
\begin{table}[t]
\caption{A comparison of CFIL to some alternatives, ablating its squasher, coupling and inductive bias. The score is the average normalized asymptotic reward over $25$ seeds ($5$ for each environment). Note: rows with two values indicate runs with smoothing of $0$ (left) and $0.5$ (right).}
\label{table:ablation_regnet}
\vskip 0.15in
\begin{center}
\begin{small}
\begin{sc}
\begin{tabular}{lcccr}
\toprule
 & score \\
\midrule
Expert & 1 \\
\midrule
CFIL    & $\textbf{1.012}$ \\
NoSquash & $-0.091$\\
RegularNet & $0.196 \mid 0.190$\\
IndFlow    & $0.158 \mid 0.127 $\\
IndFlowNS  & $0.090 \mid 0.072$\\
Numerator  & $-0.051 \mid -0.001$\\
\bottomrule
\end{tabular}
\end{sc}
\end{small}
\end{center}
\vskip -0.1in
\end{table}

\begin{figure}[t]
\vskip 0.1in
\centering
\includegraphics[width=0.84\columnwidth]{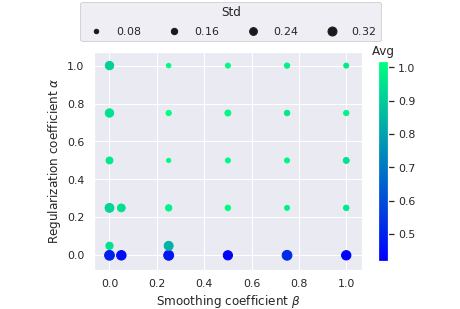} % Reduce the figure size so that it is slightly narrower than the column. Don't use precise values for figure width.This setup will avoid overfull boxes.
\caption{Averages (as color) and standard deviations (as size) of normalized asymptotic rewards for CFIL with varied levels of smoothing and regularization. Each point summarizes 25 seeds (5 per environment). The utility of the smoothing and regularization is apparent as well as CFIL's lack of sensitivity to them.}
\label{fig:ablation_smooth}
\vskip -0.1in
\end{figure}

Next we vary CFIL's smoothing and regularization coefficients to test its sensitivity. Specifically, we run CFIL (that of the three settings in Figure \ref{fig:CFILcompare}), but with pairs $\alpha, \beta \in \{0, 0.25, 0.5, 0.75, 1\},$ along with select others, and compute their averages and standard deviations of normalized asymptotic rewards over all $25$ seeds. Figure \ref{fig:ablation_smooth} illustrates the results for these runs, showcasing both the utility of the smoothing and regularization as well as CFIL's robustness to them.

%Results per environment can be found in appendix \ref{appendix:ablation_per_env}.

%% file: sections/conclusion.tex
\section{Conclusion}
We presented CFIL, a unique approach to imitation learning based on the coupling of a pair of flows. CFIL introduced many novelties including its estimator for the log ratio, its smoothing and regularization and more generally its employment of flows, while we also performed a unique analysis using BC graphs and raised concerns about trends in relevant literature. Crucially, CFIL was empirically shown to outperform state-of-the-art baselines in a variety of settings with only a single expert trajectory. A future work could include coupled flows for general ratio estimation.

%% file: sections/appendix.tex
\input{sections/appendix_sections/reproducibility.tex}

\input{sections/appendix_sections/regular_net_ablation.tex}

\input{sections/appendix_sections/justification_of_undiscounted.tex}

\input{sections/appendix_sections/normalizing_flows_appendix.tex}

\input{sections/appendix_sections/cfil_varying_traj}

\input{sections/appendix_sections/bc_extras.tex}

%% file: sections/appendix_sections/reproducibility.tex
\section{Reproducibility}

Code for reproducibility of CFIL, including a detailed description for reproducing our environment, is available at \url{https://github.com/gfreund123/cfil}.

%% file: sections/appendix_sections/regular_net_ablation.tex
\section{Ablation Results per Environment}\label{appendix:ablation_per_env}

\begin{table}[h]
\caption{Extended version of Table \ref{table:ablation_regnet}, showing the ablation results per environment. The table contains averages and standard deviations of normalized asymptotic rewards over 5 seeds.}
\label{table:ablation_full_table}
\vskip 0.15in
\begin{center}
\begin{small}
\begin{sc}
\begin{tabular}{lrrrrr}
\toprule
{} & HalfCheetah-v2 & Walker2d-v2 & Ant-v2 & Hopper-v2 & Humanoid-v2 \\
\midrule
CFIL & 1.122$\pm$0.015 & 0.8542$\pm$0.063 & 1.095$\pm$0.022 & 0.986$\pm$0.011 & 1.004$\pm$0.015 \\
NoSquash & -0.056$\pm$0.074 & -0.002$\pm$0.001 & -0.411$\pm$0.377 & 0.001$\pm$0.000 & 0.010$\pm$0.006 \\
RegularNet (smooth=0) & -0.000$\pm$0.000 & 0.149$\pm$0.002 & 0.185$\pm$0.016 & 0.291$\pm$0.009 & 0.357$\pm$0.312 \\
RegularNet (smooth=0.5)& -0.000$\pm$0.000 & 0.152$\pm$0.001 & 0.169$\pm$0.017 & 0.289$\pm$0.002 & 0.340$\pm$0.155 \\
IndFlow (smooth=0) & 0.642$\pm$0.094 & 0.000$\pm$0.002 & 0.000$\pm$0.001 & 0.138$\pm$0.306 & 0.012$\pm$0.000 \\
IndFlow (smooth=0.5) & 0.611$\pm$0.068 & 0.011$\pm$0.018 & 0.000$\pm$0.001 & 0.001$\pm$0.000 & 0.012$\pm$0.000 \\
IndFlowNS (smooth=0) & 0.203$\pm$0.183 & 0.082$\pm$0.058 & 0.126$\pm$0.176 & 0.026$\pm$0.031 & 0.017$\pm$0.001 \\
IndFlowNS (smooth=0.5) & 0.290$\pm$0.255 & 0.014$\pm$0.025 & 0.030$\pm$0.029 & 0.010$\pm$0.017 & 0.017$\pm$0.001 \\
Numerator (smooth=0) & -0.047$\pm$0.050 & 0.026$\pm$0.034 & -0.270$\pm$0.230 & 0.011$\pm$0.009 & 0.023$\pm$0.015 \\
Numerator (smooth=0.5) & -0.006$\pm$0.006 & 0.007$\pm$0.015 & -0.037$\pm$0.035 & 0.011$\pm$0.000 & 0.015$\pm$0.001 \\
\bottomrule
\end{tabular}
\end{sc}
\end{small}
\end{center}
\vskip -0.1in
\end{table}

\begin{figure}[h]
\vskip 0.1in
\centering
\includegraphics[width=0.9\columnwidth]{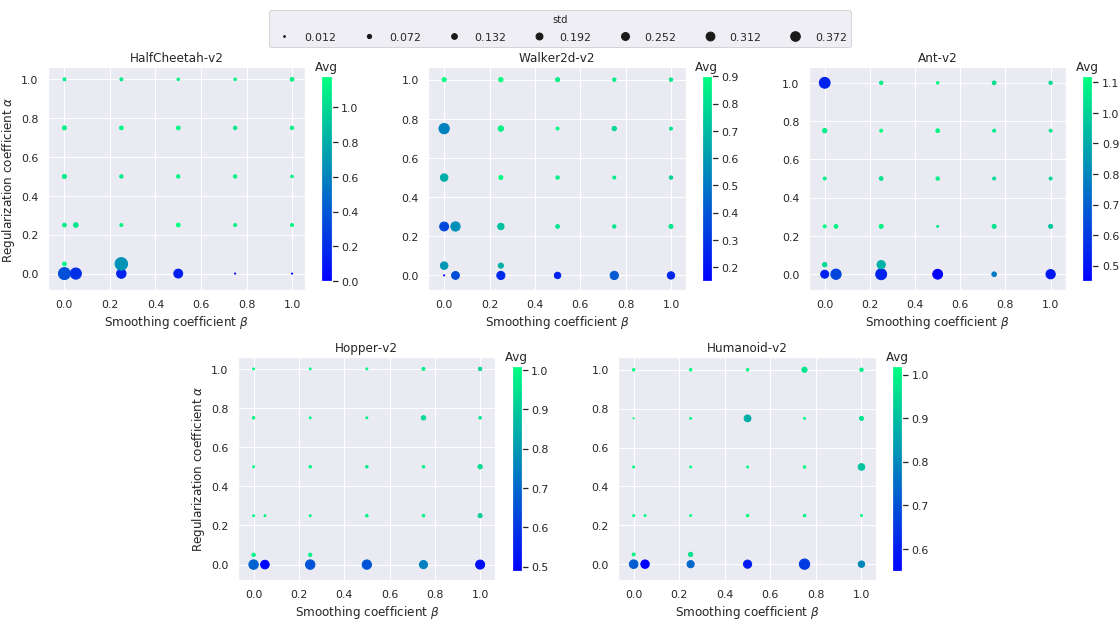} % Reduce the figure size so that it is slightly narrower than the column. Don't use precise values for figure width.This setup will avoid overfull boxes.
\caption{Extended version of Figure \ref{fig:ablation_smooth}, showing per environment the average (as color) and standard deviations (as size) of normalized asymptotic rewards for CFIL with
varied levels of smoothing and regularization. Each point summarizes 5 seeds. Once again, the importance of the smoothing and regularization is apparent, as well as CFIL lack of sensitivity.}
\label{fig:ablation_smooth_per_env}
\vskip -0.1in
\end{figure}

%% file: sections/appendix_sections/justification_of_undiscounted.tex
\section{Justification of Modeling the Undiscounted State Distribution}\label{appendix:undiscounted}
With the goal in an MDP being policy optimization, standard RL taxonomy divides between value-based and policy-based methods. The policy gradient theorem \cite{sutton2000policy}, being the foundation of policy-based methods, provides the following expression for the gradient of $J$ with respect to a parameterized policy $\pi_\theta$: 
\begin{equation}\label{eq:PGT}
\nabla_\theta J(\pi_\theta) = \sum_s d^\gamma_{\pi}(s) \sum_a \frac{d\pi_\theta(a|s)}{d\theta} Q^\pi(s,a)
\end{equation}
where $d^\gamma_{\pi}(s) = \sum_{t=0}^{\infty}\gamma^t Pr(s_t = s | s_0 \sim p_0)$ is the improper discounted state distribution. Equation \eqref{eq:PGT} enables the construction of policy optimization algorithms such as REINFORCE \cite{sutton2018reinforcement}, which are based on stochastic gradient descent (SGD). However, garnering some criticism \cite{nota2019policy}, most modern policy gradient methods \cite{schulman2017proximal,fujimoto2018addressing,haarnoja2018soft} opt to ignore the discount factor in $d^\gamma_\pi(s)$, which disqualifies them as SGD methods, hence losing guarantees it provides. Instead, they update the policy with a similar expression to \eqref{eq:PGT} in which $d^\gamma_\pi(s)$ is exchanged with the undiscounted version $d_{\pi}(s) = \sum_{t=0}^{\infty}Pr(s_t = s | s_0 \sim p_0).$  The reason being due to stability and reduction of variance (see Appendix A of \cite{haarnoja2018soft}). We in this work follow suit, modeling the undiscounted $d_{\pi}(s)$.

%% file: sections/appendix_sections/normalizing_flows_appendix.tex
\section{More on RealNVP and MAF}\label{appendix:flows}
In pursuit of inspiring more research incorporating flow-based models of the state distribution, 
%all across the literature surrounding reinforcement learning
and since flows are presumably the topic least familiar to an IL researcher reading the paper, we now provide a brief but thorough review of RealNVP and MAF, along with some additional discussion.

Following the description in Section \ref{background:flows}: For both training and generation to be efficient, $f_{\theta}$ must be easy to evaluate, easy to invert and the determinant of its Jacobian must be easy to compute. Moreover, for training to be plausible, $f_{\theta}$ must be expressive enough to capture complex high-dimensional distributions. 

RealNVP \cite{dinh2016density} uses coupling layers (distinct from our coupled approach) to achieve the above requirements. As described in the paper \cite{dinh2016density}, a coupling layer receives a $D$ dimensional input vector $x$ and outputs: 
%$y_{1:d} = x_{1:d}$ and $y_{d+1:D} = x_{d+1:D} \odot \exp(s(x_{1:d})) + t(x_{1:d})$
\begin{gather}\label{eq:Coupling} 
 y_{1:d} = x_{1:d} \\
 y_{d+1:D} = x_{d+1:D} \odot \exp(s(x_{1:d})) + t(x_{1:d}) \nonumber
\end{gather}
where $d<D$, $s$ and $t$ are functions from $R^d \to R^{D-d}$ and $\odot$ represents an element-wise product. Since coupling layers leave certain variables unchanged, they are composed in an alternating pattern to allow all variables to be transformed. Note that the Jacobian of this transformation is triangular, and its determinant is simply $exp(\sum_j{s(x_{1:d})_j)}$. Note further that the inverse of this transformation is also a simple computation: 
\begin{gather}
 x_{1:d} = y_{1:d} \\
 x_{d+1:D} = (y_{d+1:D} - t(x_{1:d})) \odot \exp( - s(x_{1:d})) \nonumber
\end{gather}
Finally, note that $s$ and $t$ have no requirement to be invertible, so they can be arbitrary neural networks. 

Masked Autoregressive Flow (MAF) \cite{papamakarios2017masked} is a generalization of RealNVP which substitutes \eqref{eq:Coupling} with the autoregressive
\begin{equation}
    y_{i} = x_{i} \exp(s_i(x_{1:i-1})) + t_i(x_{1:i-1})
\end{equation}
As stated in \ref{background:flows}, this improves expressiveness but prevents single pass sampling. However, density evaluation—our main concern—can still be performed efficiently using masked autoencoders \cite{germain2015made}.
%should verify again that i correctly described MAF

Generally, normalizing flows are sensitive objects, after all, density estimation is a difficult task, particularly with such limited data of such high dimension. Given the delicacy involved in their training, much deliberation was done over the potential influence of various things. In particular, the need for normalization, something common in other IL works \cite{kostrikov2018discriminator,kostrikov2019imitation,schroecker2020universal}. Although we did finally settle on the more robust approach of using no normalization, or any other special pre-processing, still, this could be useful and even necessary for other applications incorporating density models of the state distribution. Moreover, despite the existence of higher capacity flows \cite{kingma2018glow,huang2018neural,durkan2019neural,ho2019flow++}, when learning from such few demonstrations, seemingly more powerful flows may be far more prone to overfitting then their less expressive counterparts. We aim to inspire others to take advantage of modern powerful density estimators for explicit modeling of the state and state-action distributions, and apply them all around the reinforcement learning literature.

%% file: sections/appendix_sections/cfil_varying_traj.tex
\section{CFIL With More Trajectories}\label{appendix:CFIL_varying_traj}

%H helped figure be in the section
\begin{figure}[H]
\vskip 0.1in
\centering
\includegraphics[width=0.9\textwidth]{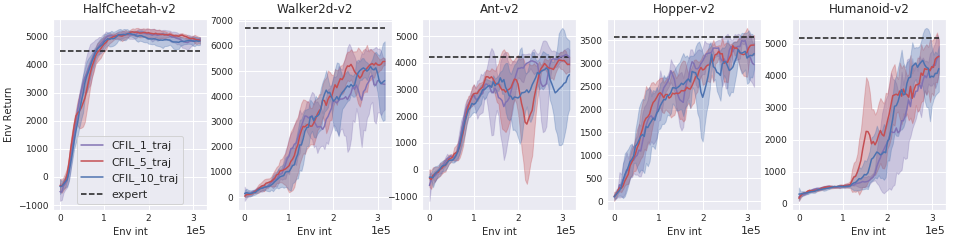}
\caption{CFIL's performance with varying numbers of expert trajectories (1, 5, and 10), demonstrating its consistency. Note, the single trajectory run is identical to that of Figure \ref{fig:CFILcompare}.}
\label{fig:CFIL_more_traj}
\vskip -0.1in
\end{figure}

%% file: sections/appendix_sections/bc_extras.tex
\section{BC Analysis Extras}

\subsection{RegularNet's BC Graph}\label{section:appendix_regular_net_bc}

\begin{figure}[h!]
\vskip 0.1in
\centering
\includegraphics[width=0.35\textwidth]{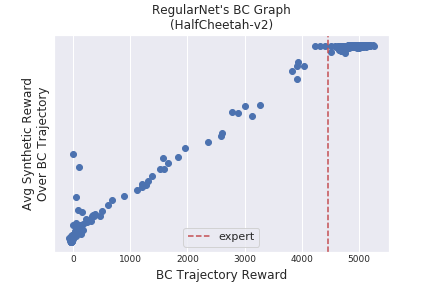} % Reduce the figure size so that it is slightly narrower than the column. Don't use precise values for figure width.This setup will avoid overfull boxes.
\caption{The BC graph of RegularNet (see ablation) for the HalfCheetah-v2 environment.}
\label{fig:regularnet_bc}
\vskip -0.1in
\end{figure}

Figure \ref{fig:regularnet_bc} shows the BC graph of RegularNet (see ablation) for the HalfCheetah-v2 environment. This serves as an example estimator which fails in RL despite a good BC graph (its failure in RL is illustrated in Table \ref{table:ablation_regnet}). As described in Section \ref{sec:our_approach}, a BC graph's utility is mainly one way. That is, when poor it is indicative of failure in RL, but the reverse is not necessarily the case. Beyond that however, it also provides insight into what is captured by an estimator as well as being valuable as a quick sanity test of an estimator’s quality (can be generated quickly on saved BC trajectories). 

\subsection{Two-Dimensional BC Analysis}
\begin{figure}[h]
\vskip 0.1in
\centering
\includegraphics[width=0.7\textwidth]{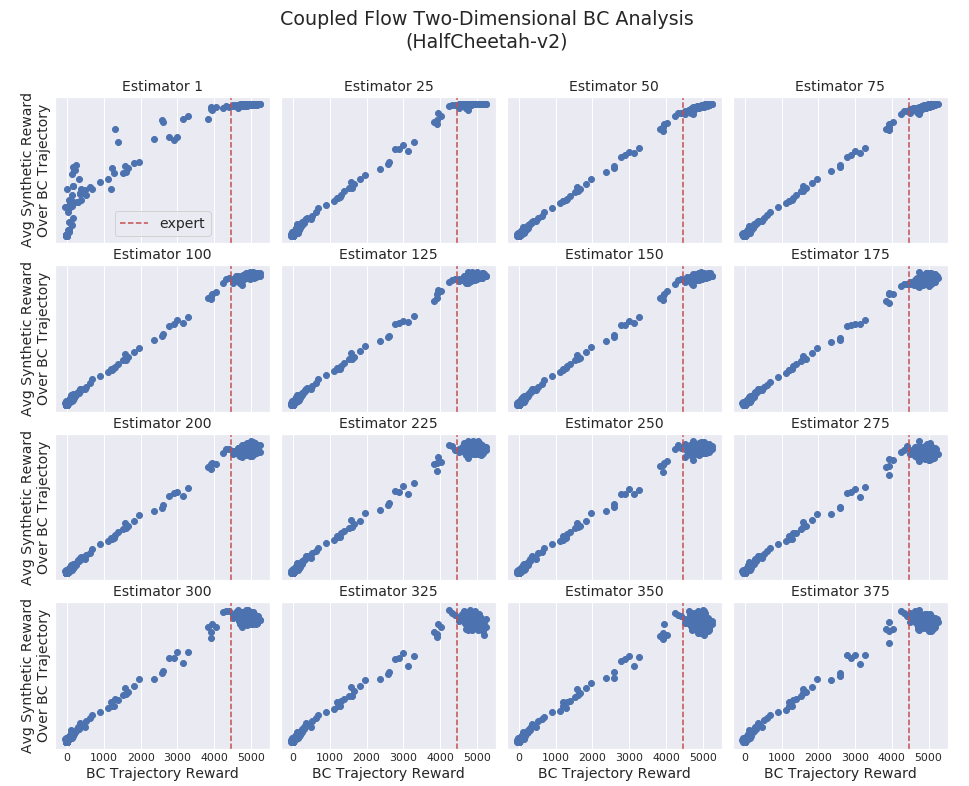} % Reduce the figure size so that it is slightly narrower than the column. Don't use precise values for figure width.This setup will avoid overfull boxes.
\caption{Two-dimensional BC analysis of coupled flows for the HalfCheetah-v2 environment. In contrast to the BC graph which is the $i$'th estimator's evaluation of the $i$'th BC trajectory vs the true environment reward of that same trajectory, here we see scatters for individual estimators along the way. That is, their evaluations of all BC trajectories scattered against the true environment rewards of the trajectories.}
\label{fig:bc_2d_coupled}
\vskip -0.1in
\end{figure}

\begin{figure}[h]
\vskip 0.1in
\centering
\includegraphics[width=0.7\textwidth]{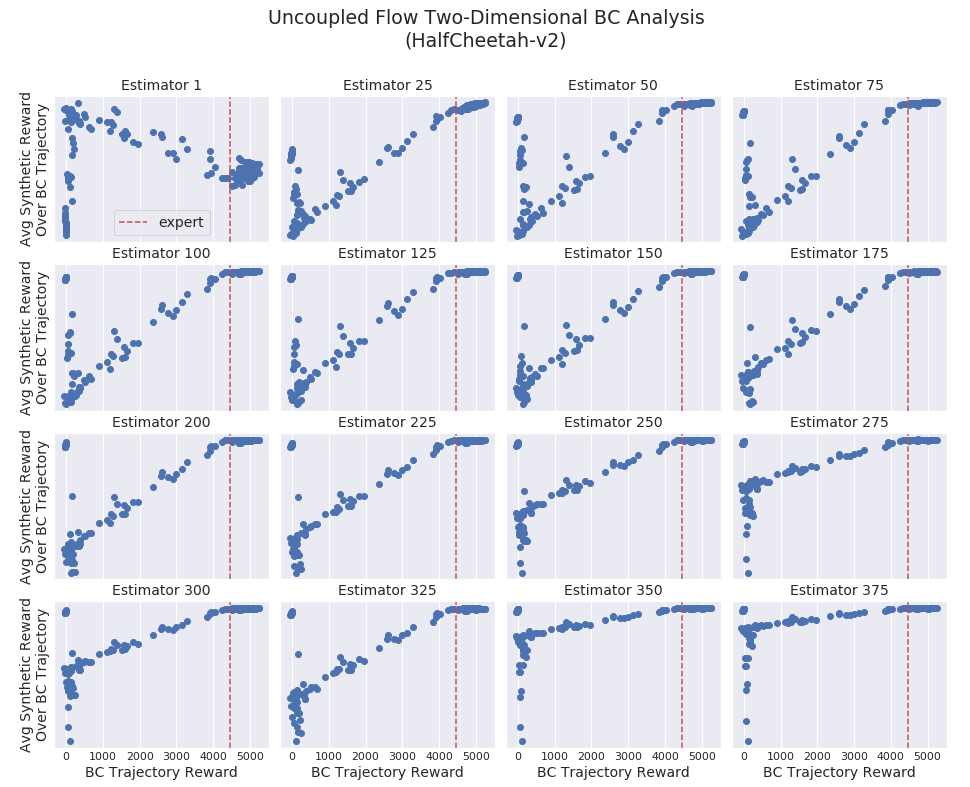} % Reduce the figure size so that it is slightly narrower than the column. Don't use precise values for figure width.This setup will avoid overfull boxes.
\caption{Two-dimensional BC analysis of uncoupled flows for the HalfCheetah-v2 environment. See Figure \ref{fig:bc_2d_coupled}'s caption.}
\label{fig:bc_2d_uncoupled}
\vskip -0.1in
\end{figure}

\begin{figure}[h]
\vskip 0.1in
\centering
\includegraphics[width=0.7\textwidth]{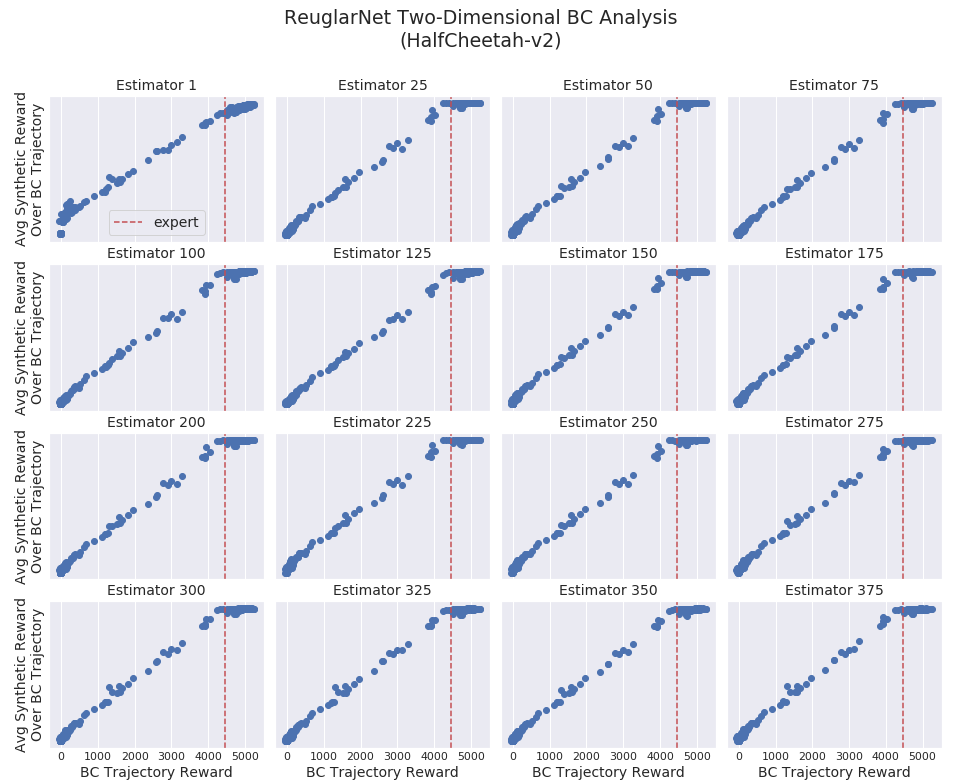} % Reduce the figure size so that it is slightly narrower than the column. Don't use precise values for figure width.This setup will avoid overfull boxes.
\caption{Two-dimensional BC analysis of RegularNet for the HalfCheetah-v2 environment. See Figure \ref{fig:bc_2d_coupled}'s caption.}
\label{fig:bc_2d_regularnet}
\vskip -0.1in
\end{figure}

What we have termed the BC graph is the diagonal slice of a broader two-dimensional picture. Recall the generation process described in Section \ref{sec:our_approach}: BC graphs show the $i$'th estimator's evaluation of the $i$'th BC trajectory against the true environment reward of that same trajectory. However, one may take interest in evaluating all $N$ estimators on all $N$ trajectories. To that end, out of the $N=375$ estimator updates, we show every $25$'th (along with the first), evaluated on all the BC trajectories, scattered against their true environment rewards. This is illustrated in Figures \ref{fig:bc_2d_coupled}, \ref{fig:bc_2d_uncoupled} and \ref{fig:bc_2d_regularnet} corresponding to coupled flows, uncoupled (independent) flows and RegularNet, respectively. The figures provide insight into what individual estimators captured as well as how they morphed over time. For example, for the coupled flows, we can see how the later estimators (synthetic rewards) incentivize the agent not to go beyond expert level, while the early ones do not. This is reasonable, since estimator $i$ is only encountered by policy $\pi_i$ (hence why the BC graph is only the diagonal cut). For this reason, evaluations of early estimators on much later trajectories aren't particularly insightful with regards to what an RL agent with the analogously constructed reward will face. Still, they're interesting in their own right, showing what has been captured by the estimator at that time.

